\documentclass[journal]{IEEEtran}
\usepackage[utf8]{inputenc}
\usepackage{hyperref}
\usepackage{authblk}
\usepackage{graphicx}
\usepackage{caption}
\usepackage{bm}
\usepackage{color}
\usepackage{amsmath}
\usepackage{amssymb}
\usepackage{graphicx}
\usepackage{multirow}

\def\btheta{{\boldsymbol{\theta}}}

\def\bphi{{\boldsymbol{\phi}}}

\newcommand{\outNim}[1]{}
\DeclareMathOperator*{\argmax}{arg\,max}
\DeclareMathOperator*{\argmin}{arg\,min}
\usepackage[ruled,vlined]{algorithm2e}
\usepackage{gensymb}
\usepackage{subcaption}
\usepackage{amssymb}

\begin{document}

\title{Examining the Benefits of Capsule Neural Networks}
\author[1*]{Arjun Punjabi}
\author[2*]{Jonas Schmid}
\author[1]{Aggelos K. Katsaggelos}

\affil[1]{Electrical and Computer Engineering, McCormick School of Engineering, Northwestern University}
\affil[2]{Electrical Engineering, Institute for Communication Systems, HSR University of Applied Science Rapperswil}
\affil[*]{Both authors contributed equally to this work}
\date{May 2019}

\maketitle

\begin{abstract}
Capsule networks are a recently developed class of neural networks that potentially address some of the deficiencies with traditional convolutional neural networks. By replacing the standard scalar activations with vectors, and by connecting the artificial neurons in a new way, capsule networks aim to be the next great development for computer vision applications. However, in order to determine whether these networks truly operate differently than traditional networks, one must look at the differences in the capsule features. To this end, we perform several analyses with the purpose of elucidating capsule features and determining whether they perform as described in the initial publication. First, we perform a deep visualization analysis to visually compare capsule features and convolutional neural network features. Then, we look at the ability for capsule features to encode information across the vector components and address what changes in the capsule architecture provides the most benefit. Finally, we look at how well the capsule features are able to encode instantiation parameters of class objects via visual transformations.
\end{abstract}


\section{Introduction}
\IEEEPARstart{C}{onvolutional} neural networks (CNNs) have long been the tools of choice when tackling computer vision problems. The spatial localization of CNN features is greatly beneficial when the networks are applied to images and videos; however, these networks also have their shortcomings. The kernels in a convolutional layer must learn to identify the presence of all relevant features in the input. Thus, transformations such as rotations and occlusion can be detrimental when the training dataset is not properly augmented. Even still, the burden of learning visual features in addition to all possible modifications of these features can be immense for a traditional CNN. 

Recently, a novel class of neural networks was proposed in \cite{sabour2017dynamic} that employs the concept of a "capsule". The authors describe a capsule as a group of neurons that represent the existence of a feature in addition to parameters regarding the instantiation of said feature. Contrary to the scalar activations of kernels in a traditional CNN, these capsule vectors aim to be richer representations of information in the network. In this manner, a capsule should be able to encode not only the existence of a particular visual feature, but also the transformations it can undergo in the given application.

That said, while initial results show great potential for capsule networks, there is still much uncertainty regarding how these capsules function. In fact, the "black box" analogy can be applied to all classes of neural networks, not just those with capsules. The interpretability of neural networks has always been a problem, and it is difficult to examine the benefit of capsules without a comparison to traditional CNN features.

In an attempt to elucidate these capsules, this investigation will begin by employing a deep visualization technique to generate images that visually represent the information contained in a capsule. This image can then be compared to an image created in a similar fashion from a traditional CNN, and the discrepancies between them can provide visual justification for the hypothesized benefits of capsule networks. Furthermore, the visual impact of modifying values in a capsule are examined to more accurately ascertain their capacity. Finally, the investigation will examine other facets of the original capsule network architecture proposed in \cite{sabour2017dynamic}, namely the benefits of dynamic routing and a reconstruction network. The next section highlights related work in the field, followed by an outline of the capsule network and visualization methodologies. Finally, the results are shown and the resulting trends are discussed.

\section{Related Work}

The concept of a capsule neural network originated in Hinton's 2017 paper \cite{sabour2017dynamic}, wherein the capsule vectors are described and implemented within a convolutional architecture. Furthermore, a dynamic routing algorithm is proposed that selectively links units in a capsule together rather than traditional downsampling methods such as max pooling. There is a follow up publication from Hinton in 2018 \cite{hinton2018matrix} that extends capsules to matrix form as well as further developing the routing scheme; however, our work will primarily focus on the architecture discussed in \cite{sabour2017dynamic}, and our experiments will be in parallel to those performed in the first publication.

Other modifications to the original architecture have also been proposed, including in \cite{xi2017capsule} where the capacity of the network is increased (both via numbers of layers and size of capsules) along with changes to the activation function. The authors in \cite{rawlinson2018sparse} demonstrate that capsules without the masking operation used in \cite{sabour2017dynamic} may generalize better. The work in \cite{xiang2018ms} extends the capsule scheme to a multi-scale hierarchy. A generative adversarial network (GAN) is proposed in \cite{jaiswal2018capsulegan} that makes use of capsules in the discriminator network. The network in \cite{deng2018hyperspectral} takes hyperspectral images as input as opposed to standard RGB images. The authors in \cite{neill2018siamese} create a Siamese capsule network by combining pairwise inputs with the capsule architecture.

The applications for capsule networks have also been widespread. In \cite{li2018recognition}, a capsule network uses images taken by a UAV for classification of rice fields. A detection problem is performed to find street signs in \cite{kumar2018novel}, while the authors in \cite{kim2018capsule} use capsules to analyze traffic patterns in a city. The work in \cite{islam2018seagrass} outlines a capsule network for seagrass segmentation in satellite images. Video data is used as input to an action detection network using capsules in \cite{duarte2018videocapsulenet}. Capsules have also been used on text data for classification \cite{ren2018compositional} and sentiment analysis \cite{wang2018sentiment}. The authors in \cite{andersen2018deep} design a reinforcement learning approach with capsules to play complex games.

Despite their relative nascency, many have started using capsules in the medical domain, including for segmentation \cite{lalonde2018capsules} and cancer detection \cite{mobiny2018fast} in lung CT scans. These networks have also been used on MRI data for brain tumor classification \cite{afshar2018brain} and histology images for breast cancer identification \cite{iesmantas2018convolutional}. The authors in \cite{jimenez2018capsule} discuss challenges of using public medical datasets in the context of capsule networks. Finally, \cite{bahadori2018spectral} proposes a spectral capsule network to solve the "learning to diagnose" problem.

Clearly, these capsule networks exhibit great potential; yet, the justification for how these networks perform so well is less clear. Granted, Hinton enumerates several potential benefits of capsules in \cite{sabour2017dynamic}, namely that the increased dimensionality of the capsules allows feature transformation encoding and that dynamic routing is a more intelligent way of aggregating information. That said, the experimental results, while impressive, are not necessarily proof of that the capsules are exhibiting these traits. One set of experiments in \cite{sabour2017dynamic} seems to indicate that certain object features can be controlled via capsule manipulation, but this is not explored to greater depth. The authors in \cite{shahroudnejad2018improved} make a more concerted attempt at explainability by varying output capsules in more than one dimension, but yet again this methodology is somewhat limited in scope.

Our investigation compounds on these capsule manipulation experiments by adding deep visualization techniques. These techniques are aimed at creating images from a trained network that represent the information contained in the weights. From this, one can gain greater insight into how a model functions and what features are in use. The primary technique we employ is activation maximization, which is described in \cite{mahendran2015understanding, erhan2009visualizing} and generalized in \cite{mahendran2016visualizing} to describe other techniques, including inversion \cite{dosovitskiy2016inverting}. While we will not extend much beyond the activation maximization structure, other investigations have looked at CNN features with attention maps \cite{yosinski2015understanding, zeiler2014visualizing} and saliency maps \cite{simonyan2013deep}. The authors in \cite{nguyen2016multifaceted} use activation maximization and a clustering approach to visualize different "facets" of image classes, the visualizations are performed while training in \cite{punjabi2017visualization}, and the Google deep dream generator \cite{alexander2015inceptionism} uses a procedure similar to activation maximization to create art.

Fundamentally, the application of activation maximization to a capsule network for the purpose of understanding the benefits over a traditional CNN is a nascent investigation. Moreover, given that the justification for capsule networks at a feature level has not been thoroughly explored, the necessity for understanding capsules before adopting them in the field is paramount. In the next section, we describe activation maximization and the other methods we employ for the purpose of analyzing capsule features.

\section{Methodology}

To begin, our investigation applies the deep visualization technique of activation maximization to two trained neural networks: a capsule neural network, and a CNN with comparable computational power and information capacity. By comparing the resulting images, we are able to distinguish the different feature representations in these two networks and glean insight into the potential benefits of capsules. The second experiment further scrutinizes the capsule features in order to more directly ascertain whether capsule vectors truly model transformation parameters. This is done by applying a principal component analysis (PCA) on a set of manually transformed images. The resulting PCA spaces indicate structure in the capsule vectors related to the respective transformations. Then, to even further demonstrate the transformation encoding ability of capsules, a modified activation maximization procedure is used to generate images that correspond to said transformations. By modifying capsule vectors along the principal components to varying degrees and then using the modified activation maximization procedure, we can see that the capsules can generate images with varying degrees of visual transformations. Finally, these investigations are performed when the reconstruction network that is typically present in the capsule network architectures is removed. Some results are also shown in the case when dynamic routing is removed. This section will outline the capsule network architecture, activation maximization algorithm, and how these are used in conjunction with PCA to perform energy compaction and transformation encoding on capsule vectors. Specific results for these methods will follow after some experimental details.

\subsection{Capsule Network Architecture}

\begin{figure*}[ht]
\centering
\includegraphics[width=\linewidth]{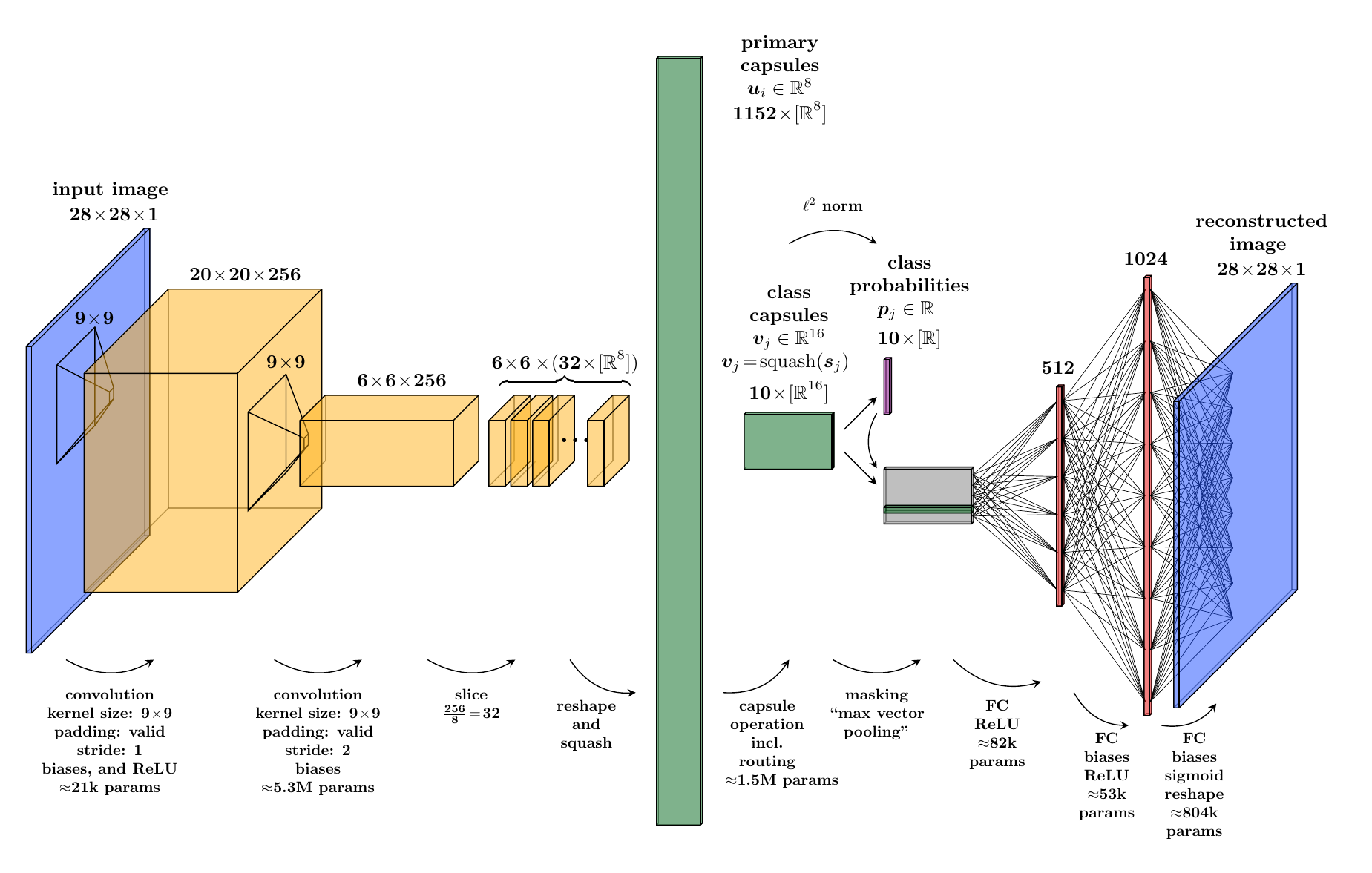}
\caption{Capsule network architecture including reconstruction network used as regularizer}
\label{fig:caps_net}
\end{figure*}

This investigation employs an architecture identical to the one outlined in \cite{sabour2017dynamic}, as shown in Fig. \ref{fig:caps_net}. The network takes as input a 28 x 28 grayscale image and proceeds with a standard convolutional layer with ReLU activation, followed by a strided convolution layer. At this point, the feature maps are split into groups before being reshaped into the primary capsule layer. The nonlinearity used for the last step is the "squash" function developed in \cite{sabour2017dynamic} and defined by:
\begin{equation}
\textbf{v}_j = \frac{\lvert \lvert \textbf{s}_j \rvert \rvert ^ 2}{1 + \lvert \lvert \textbf{s}_j \rvert \rvert ^ 2} \frac{\textbf{s}_j}{\lvert \lvert \textbf{s}_j \rvert \rvert}
\end{equation}
where $\textbf{v}_j$ $\epsilon$ $\mathbf{R}^8$ is the vector output of the capsule and $\textbf{s}_j$ is its input. This activation function aims to maintain the direction of a capsule vector while normalizing its length such that short vectors are mapped to vectors with near zero length while long vectors are mapped to vectors with length close to one. The class capsule layer follows the primary capsule layer, and it is at this point where the dynamic routing algorithm is implemented. This "routing by agreement" serves as a more advanced method of neuron connection as compared to traditional methods like max-pooling which can lose all but the most prominent connections. Again from \cite{sabour2017dynamic}, the capsule input is given by:
\begin{equation}
\textbf{s}_j = \sum_{i}c_{ij} \hat{\textbf{u}}_{j\lvert i}
\end{equation}
with
\begin{equation}
\hat{\textbf{u}}_{j\lvert i} = \textbf{W}_{ij} \textbf{u}_i
\end{equation}
where $\hat{\textbf{u}}_{j\lvert i}$ are the prediction vectors found by multiplying the capsule vectors in the previous layer $\textbf{u}_i$ by the weight matrices of the layer $\textbf{W}_{ij}$. The coupling coefficients $c_{ij}$ used in the dynamic routing process are given by the "routing softmax":
\begin{equation}
c_{ij} = \frac{\exp{b_{ij}}}{\sum_{k}\exp{b_{ik}}}
\end{equation}
where $b_{ij}$ are the logits of the coupling coefficients that are iteratively refined by the routing algorithm as proposed in \cite{sabour2017dynamic}. The initial logits are set to zero in all our experiments. In doing so, the coefficients converge towards agreement of the output of one capsule $\textbf{v}_j$ with the output of a capsule in the previous layer $\hat{\textbf{u}}_{j\lvert i}$.

\begin{algorithm}
    \SetAlgoLined
    \Begin{
    for all capsule $i$ in layer $l$ and capsule $j$ in layer $(l + 1)$: $b_{ij} \leftarrow 0$
    
    \For{$r$ iterations}{
    for all capsule $i$ in layer $l$: $\textbf{c}_i \leftarrow$ softmax($\textbf{b}_i$)
    
    for all capsule $j$ in layer $(l + 1)$: $\textbf{s}_j \leftarrow \sum_{i}c_{ij} \hat{\textbf{u}}_{j\lvert i}$
    
    for all capsule $j$ in layer $(l + 1)$: $\textbf{v}_j \leftarrow$ squash($\textbf{s}_j$)
    
    for all capsule $i$ in layer $l$ and capsule $j$ in layer $(l + 1)$: $b_{ij} \leftarrow b_{ij} + \hat{\textbf{u}}_{j\lvert i} \textbf{v}_j$
    
    }
    }
    \caption{Routing Algorithm\label{alg}}
\end{algorithm}

After the class capsules are found, the $l^2$ norm of $\textbf{v}_j$ is used to find the class probabilities, which in turn are used to make the final classification. While this is the entirety of the network at testing time, it is trained with a reconstruction network that takes the output of the largest capsule vector (corresponding to the classification label) and applies three fully connected layers. The output of these layers is the same size as the reconstructed image, and the mean squared error of this image and the input of the total network is used as an added term in the loss function. This reconstruction network acts as a method of regularization to ensure that the capsules maintain sufficient information to represent the input. With the network defined, we now describe the activation maximization method performed on a trained capsule network.

\begin{figure*}[ht]
\centering
\includegraphics[width=\linewidth]{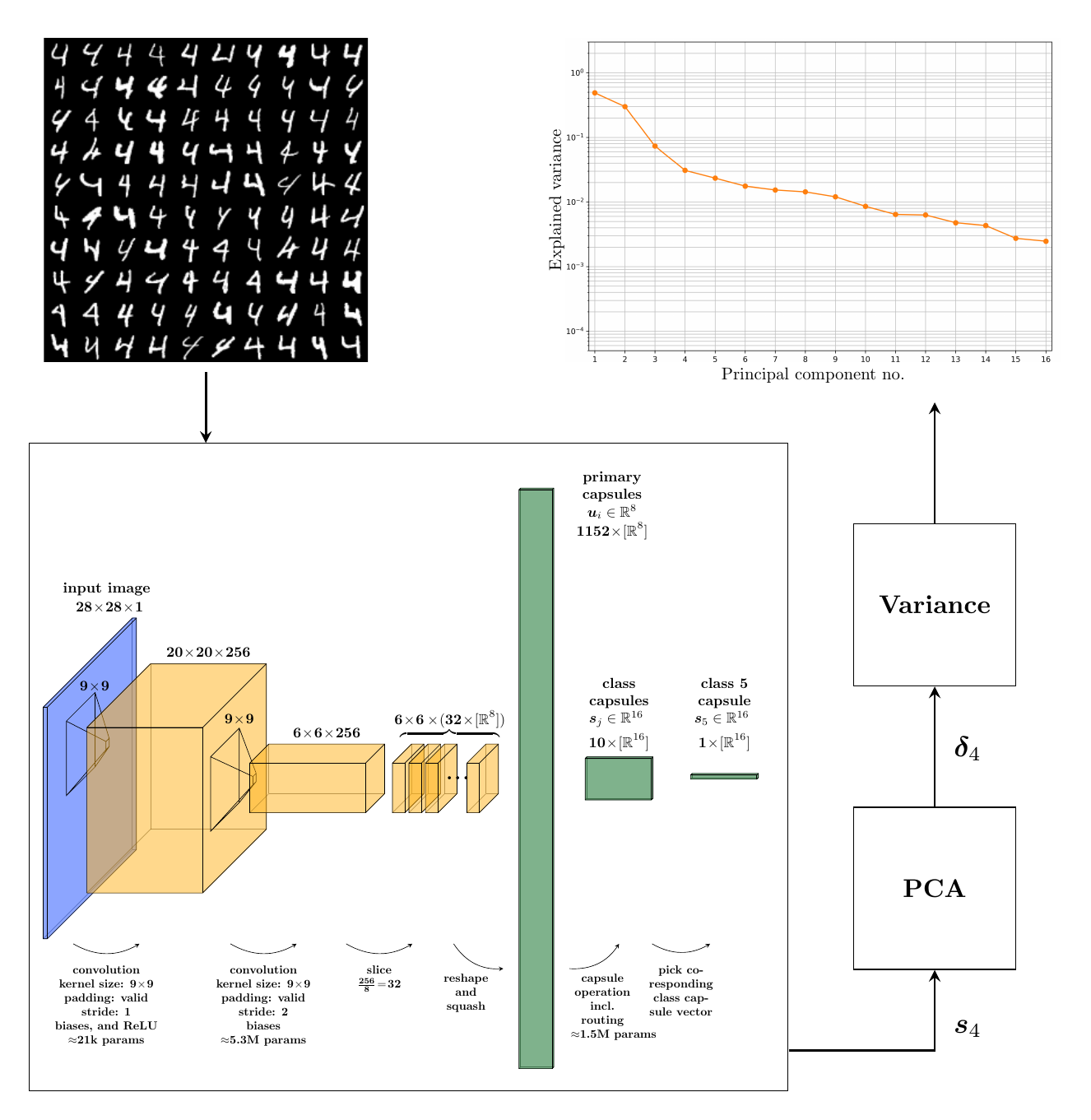}
\caption{Energy compaction for set of images}
\label{fig:nrg_comp}
\end{figure*}

\subsection{Activation Maximization}

After the network is trained, activation maximization can be used as a means of visualizing the features learned by the network. In general, activation maximization is an optimization approach to produce images that can represent either intermediate network features or object classes. It can be formulated as:
\begin{equation}
x^* = \argmax_{x} (a_i(x) - R_\btheta(x))
\end{equation}
where $x^*$ is the final visualization, $x$ is a candidate input image to the network, $a_i(x)$ is the activation for some particular unit $i$, and $R_\btheta(x)$ is some parameterized regularization function. Depending on the choice of the unit $i$, the visualization represents different kinds of information. In a CNN, if $i$ is chosen to be the index of a filter in a convolutional layer, the visualization will depict an image that corresponds to the maximum output of the filter. Depending on the choice of filter, these visualizations could manifest as object components or texture patterns. If, on the other hand, $i$ is chosen to be an element in the final layer class probability vector, the visualization will depict the aggregation of network features that most strongly represents the class. In other words, these images should be the most optimal exemplars of the class. We will be employing this functionality on both a CNN as well as on a single element in the final layer of a capsule network.

We will also use a slightly modified version of this procedure. To begin, we restructure the problem from a maximization to a minimization:
\begin{equation}
x^* = \argmin_{x} (l(\bphi(x),\bphi_0) + R_\btheta(x))
\end{equation}
where $l(\bphi(x),\bphi_0)$  is a loss function between the feature representation of the input $\bphi(x)$ and the target feature representation $\bphi_0$. If $\bphi_0$ is chosen to be a one-hot indicator vector for a given class, then the result of this optimization is the same again an exemplar image from the class. However, one can also choose a different $\bphi_0$, such as the capsule vector found by passing a particular image through the network. This would create an image that very closely resembles the input image. As such, it is more well defined as an "activation matching" procedure rather than the traditional title of activation maximization. We use this technique later when analyzing the transformation encodings.

\subsection{Energy Compaction}

While the activation maximization images provide a method to visually examine network features, the energy compaction analysis outlined here presents a more quantitative comparison. After the capsule network is trained, a set of images is passed through it in order to obtain the corresponding capsule vectors. Principal component analysis (PCA) is performed on these vectors, following which the variance along each of these particular dimensions is found. One can see a diagram of this pipeline in Fig. \ref{fig:nrg_comp}.

This variance is a measurement related to the distribution of energy or information along a specific principal axis. If the information is heavily concentrated in one particular direction in the native capsule space, the variance along the first principal component will be disproportionately large. This in turn indicates that the number of dimensions required to represent the information is small, with perhaps even just one dimension being sufficient. In this investigation, this phenomenon correlates with capsule vectors that do not encode information across all dimensions in the capsule space; rather, these vectors only encode information in a small number of dimensions. Because the capsule vectors proposed in \cite{sabour2017dynamic} are supposed to store instantiation parameters, this may indicate that the capsules are not functioning optimally. Thus, the benefit over traditional CNN features may also be limited.

The converse is also true: when the information in the capsule vector is well distributed, the variance along the principal axes will be more balanced. Granted, any PCA will yield components that have decreasing variance as the component number increases; however, the slope of this trajectory is more gradual in this case. Consequently, many if not most of the components are required to represent the information in the capsule vectors. This potentially correlates with capsule vectors that are functioning optimally and suggests a benefit over CNN features in line with the findings in \cite{sabour2017dynamic}.

\begin{figure*}[ht]
\centering
\includegraphics[width=\linewidth]{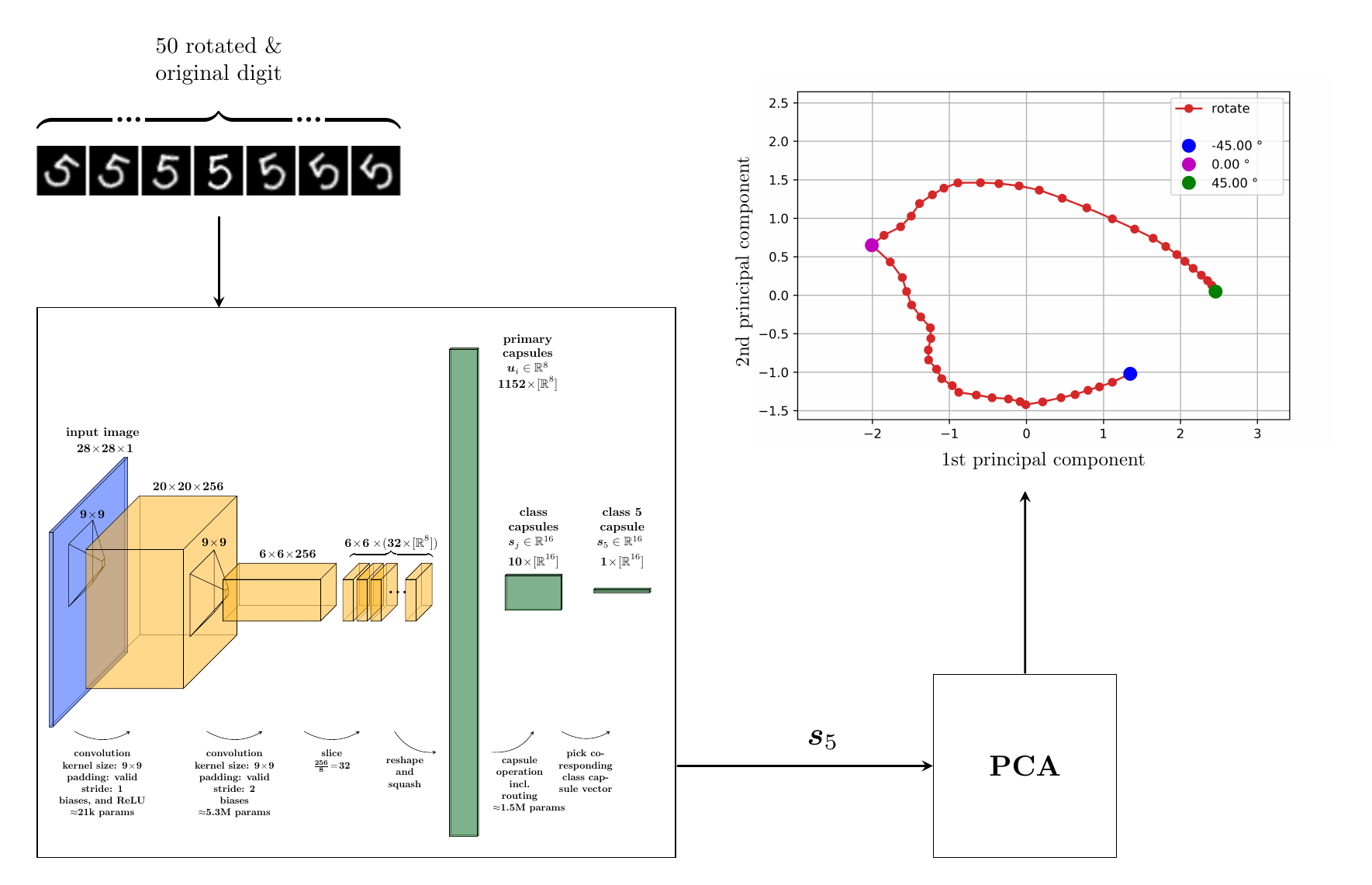}
\caption{Transformation encoding forward analysis for rotation}
\label{fig:forward_analysis}
\end{figure*}

\begin{figure*}[ht]
\centering
\includegraphics[width=\linewidth]{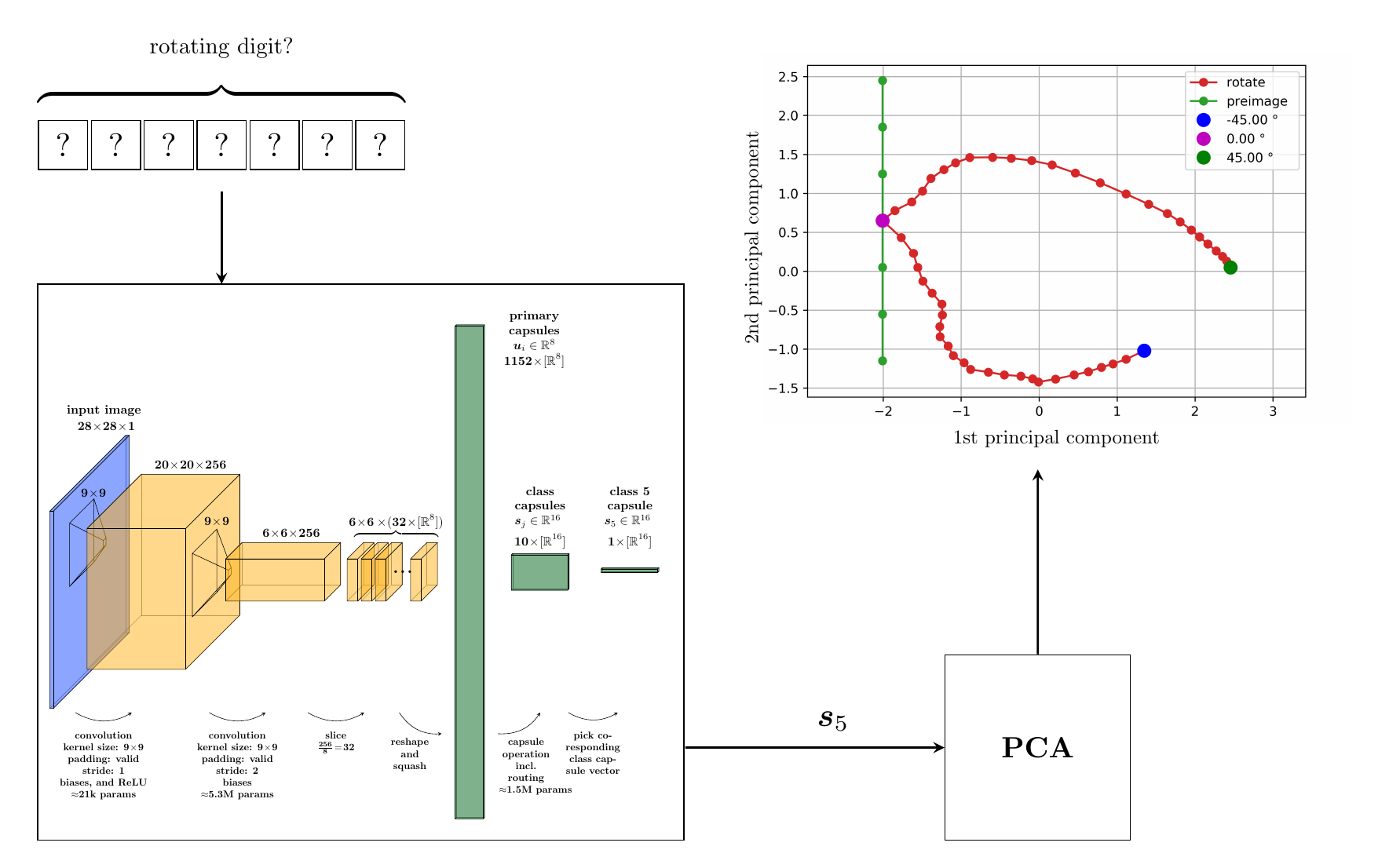}
\caption{Transformation encoding pre-image analysis on generated PCA space}
\label{fig:preimage_analysis}
\end{figure*}

\begin{figure*}[ht]
\centering
\includegraphics[width=0.7\linewidth]{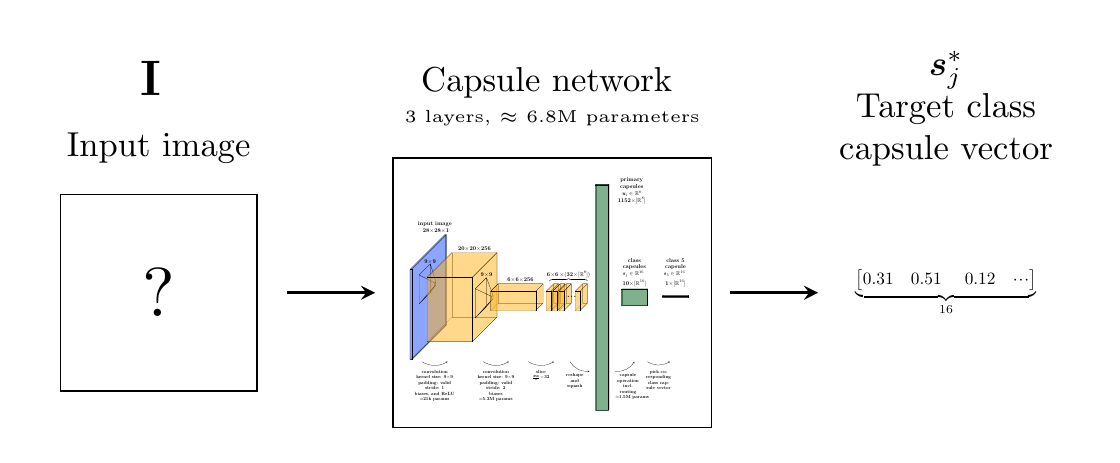}
\caption{Activation maximization/matching for pre-image analysis}
\label{fig:preimage_am}
\end{figure*}

\subsection{Transformation Encoding}

To take the energy compaction analysis one step further, we perform a transformation encoding analysis that uses the developed PCA-based framework to examine how capsule vectors encode image transformations. The authors in \cite{sabour2017dynamic} claim that capsule vectors should be able to encode image transformations such as rotation and scale changes. In order to examine this claim, we perform both a forward analysis and a pre-image analysis.

\subsubsection{Forward Analysis}

The forward analysis begins with a manually generated set of transformed images. One can see an example of this in Fig. \ref{fig:forward_analysis}, where an image from the MNIST dataset (shown in the center of the row of images) is manually transformed with varying degrees of rotation. The resulting images are used as input to the energy compaction procedure which yields principal components as before. Instead of plotting the variance as a function of principal component index, the plot in Fig. \ref{fig:forward_analysis} shows each image as a single point on a 2D grid spanned by the first and second component values. For example, the original image (purple point) has a first principal component value of approximately equal to -2 and a second principal component value of approximately equal to 0.7. The points are linked to show how the images span the principal component space with increasing transformation intensity. The green and blue points in the plot show the images with the largest transformations, which in this case are the images with 45\degree and -45\degree rotation, respectively. One can glean some insight from observing the shape of these curves; for example, a smooth transformation curve that is oriented with principal component axes is indicative of relative organization in the capsule domain. However, the more explanatory results are shown when the forward analysis is followed by pre-image analysis.

\subsubsection{Pre-image Analysis}

The term "pre-image" has been used in several ways in the deep visualization literature. Here, we use the term to define an image whose capsule vector most closely matches a particular target. Continuing the example described in Fig. \ref{fig:forward_analysis}, we see in Fig. \ref{fig:preimage_analysis} that the pre-image analysis aims to find images that match some target values when passed through the network and whose capsule vectors are then transformed into PCA space. While the objective of the forward analysis was to generate transformation curves in the PCA space, the objective of the pre-image analysis is to ascertain the ability to control visual transformation via capsule vector modification. The ability to do so further justifies the claims made in \cite{sabour2017dynamic}. In Fig. \ref{fig:preimage_analysis}, the green points represent modified capsule vectors in the PCA space. In this particular case, the original digit image without rotation (shown in purple) had the second principal component value modified with varying degrees to form a set of pre-images. A modified activation maximization procedure, perhaps more accurately described as activation "matching", was used to find the corresponding pre-images and is shown in Fig. \ref{fig:preimage_am}. This procedure is formulated in almost the same way as before; however, instead of minimizing the loss between the feature representation of the input and an indicator function, the target was chosen to be the modified class capsules vectors. After performing this optimization, one will find images similar to those used in the forward analysis. These images can then be used to ascertain the capsule vectors' robustness to image transformations.

\section{Results and Discussion}

The capsule network was trained on the MNIST dataset of handwritten digits in the same manner as \cite{sabour2017dynamic}. Therefore, the ten capsules in the final capsule layer each correspond to a particular digit. When capsule vectors need to be isolated for a particular class in either the activation maximization or PCA-based procedures, the respective row of the capsule matrix is taken for further processing. The network itself was implemented in Tensorflow and trained on a single NVIDIA Tesla P100 GPU. The Adam optimizer \cite{kingma2014adam} was used with the originally proposed decay rates and the resulting training times were approximately 15 hours when routing was used and 12 hours when routing was omitted. A baseline CNN architecture with similar computational cost was also trained on the MNIST data using the specifications outlined in \cite{sabour2017dynamic}. 

In the activation maximization and pre-image algorithms, two forms of regularization were used; first, a median filter of kernel size 3x3 was applied every 100th gradient step and second, pixel values outsize the normalized 0 to 1 range were clipped at each step. These two regularization methods ensured that the resulting images were interpretable and stayed within the distribution of the original dataset. The following sections describe and discuss the results from each of the previously formulated methods using this experimental setup.

\subsection{Comparison of Classification Methods}

\begin{table}[h!]
\centering
\begin{tabular}{|c|c|c|c|}
\hline
Model & \multicolumn{2}{|c|}{Configuration} & Error Rate (\%)\\
& Reconstruction & Routing & \\
\hline
Baseline CNN & - & - & 0.49$\pm0.027$\\
CapsNet & no & no & 0.34$\pm0.020$\\
CapsNet & weak & no & 0.33$\pm0.030$\\
CapsNet & strong & no & 0.33$\pm0.024$\\
CapsNet & no & yes & 0.39$\pm0.019$\\
CapsNet & weak & yes & 0.31$\pm0.031$\\
CapsNet & strong & yes & \textbf{0.28$\pm0.017$}\\
\hline
\end{tabular}
\caption{Classification Error Rates for Network Configurations (5 Trials)}
\label{table:classification}
\end{table}

After training, the classification error rates of the capsule network and CNN were 0.28\% and 0.49\%, respectively. In Table \ref{table:classification}, one can also see the classification performance of the capsule network configurations with varying amounts of the reconstruction regularizer and dynamic routing. The dynamic routing algorithm can simply be turned on or off. In the the latter case, the capsules are still structured as previously described; however, the routing coefficients are not iteratively modified as in the algorithm. The reconstruction configuration is defined as "no", "weak", or "strong". When the network does not use the reconstruction component, the corresponding term in the loss function is set to zero. In the "strong" case, the term weight is 20 times larger than in the "weak" case. From the table, one can see that the capsule network outperforms the baseline CNN in all cases, and furthermore the addition of strong reconstruction and routing does improve the performance. It is important to note the relative importance of each of these components. In the case when no routing is used, the reconstruction network had minimal impact on performance. When routing is used, increasing the weight of the reconstruction loss reduced the error rate. This indicates that, while the proposed dynamic routing contribution in \cite{sabour2017dynamic} does have benefits in a capsule network architecture, the relative benefit of the reconstruction network should not be understated. Without such regularization, the dynamic routing alone does not necessarily provide a benefit as it may even hinder classification performance. With all this said, one may point out that the classification margins are very slim between all these cases given that the networks all exceed 99\% classification accuracy on the MNIST testing set. As a result, other means of comparison are necessary to obtain an accurate picture of the salutary effects of capsules. With this in mind, the activation maximization results are a first step in looking more deeply at capsules.

\subsection{Activation Maximization}

\begin{figure*}[ht]
\centering
\begin{subfigure}{.45\textwidth}
    \centering
    \includegraphics[width=\linewidth]{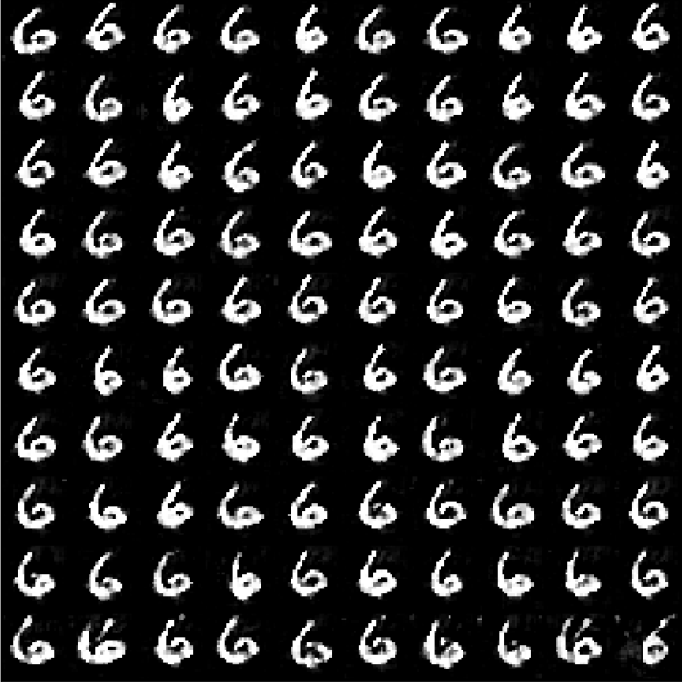}
    \caption{Capsule network, strong reconstruction}
    \label{fig:am_6_caps}
\end{subfigure}
\begin{subfigure}{.45\textwidth}
    \centering
    \includegraphics[width=\linewidth]{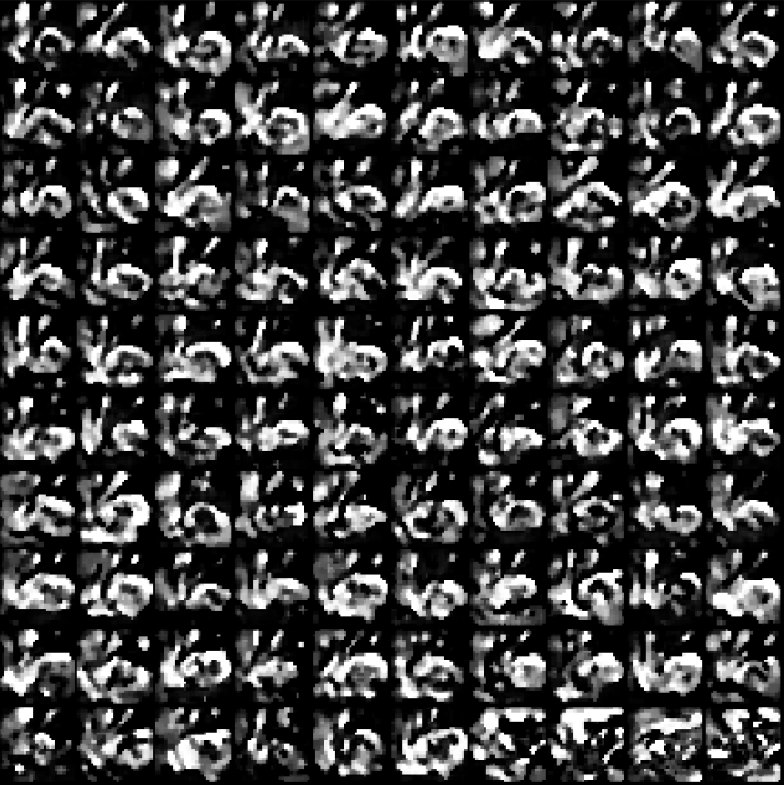}
    \caption{Capsule network, no reconstruction}
    \label{fig:am_6_caps_norecon}
\end{subfigure}
\begin{subfigure}{.45\textwidth}
    \centering
    \includegraphics[width=\linewidth]{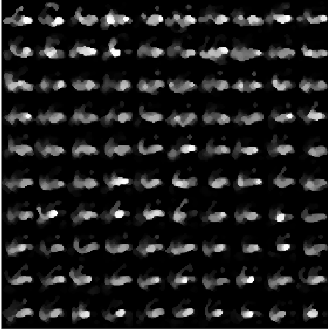}
    \caption{CNN}
    \label{fig:am_6_cnn}
\end{subfigure}
\caption{Activation maximizations for digit 6}
\label{fig:am_6}
\end{figure*}

\begin{figure*}[ht]
\centering
\begin{subfigure}{.45\textwidth}
    \centering
    \includegraphics[width=\linewidth]{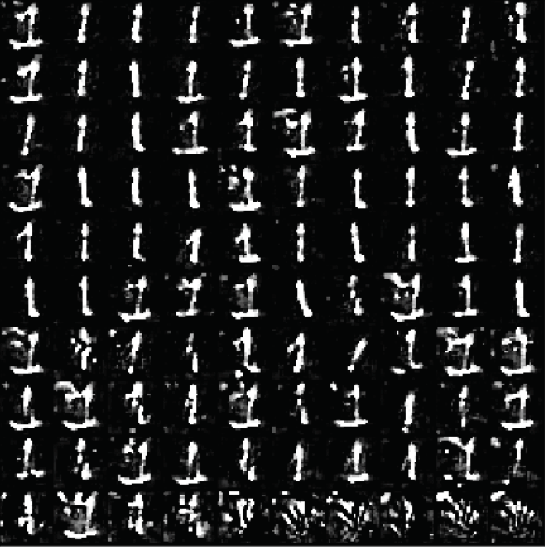}
    \caption{Capsule network, strong reconstruction}
    \label{fig:am_1_caps}
\end{subfigure}
\begin{subfigure}{.45\textwidth}
    \centering
    \includegraphics[width=\linewidth]{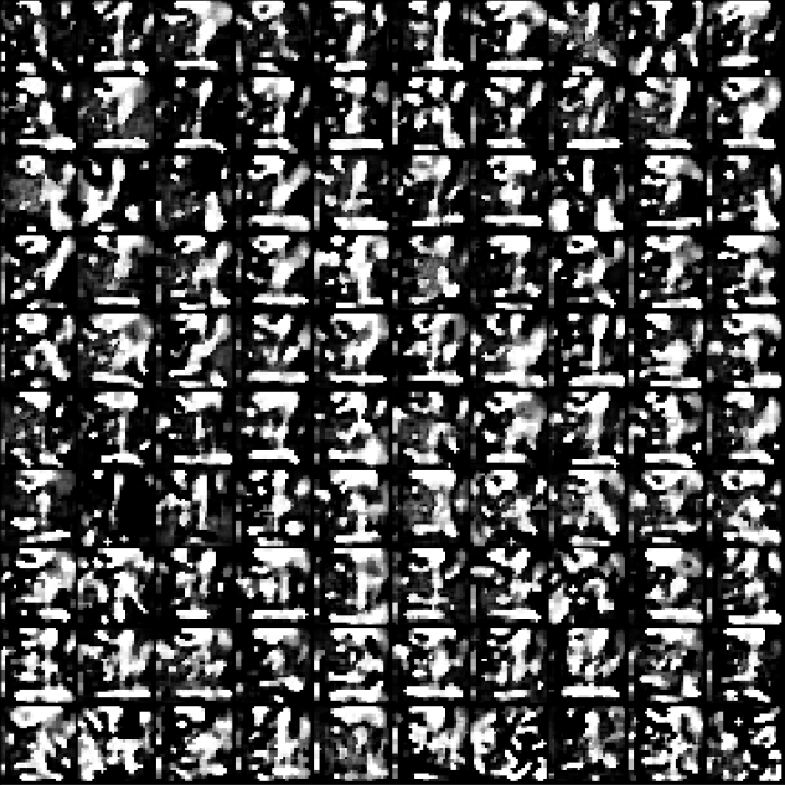}
    \caption{Capsule network, no reconstruction}
    \label{fig:am_1_caps_norecon}
\end{subfigure}
\begin{subfigure}{.45\textwidth}
    \centering
    \includegraphics[width=\linewidth]{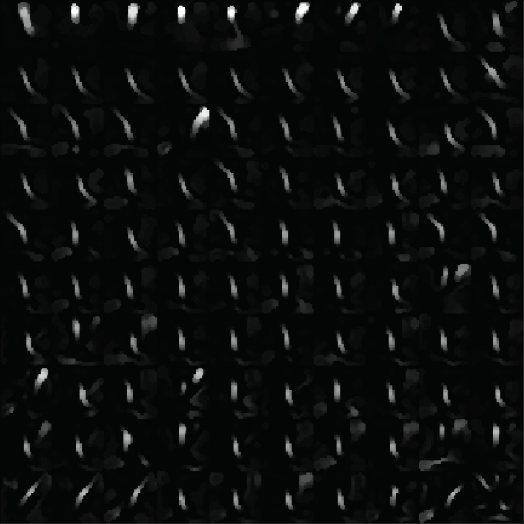}
    \caption{CNN}
    \label{fig:am_1_cnn}
\end{subfigure}
\caption{Activation maximizations for digit 1}
\label{fig:am_1}
\end{figure*}

As discussed, the activation maximization analysis aims to create images that represent information learned by a network. Fig. \ref{fig:am_6} shows 100 such images created from the activation maximization algorithm when applied to a capsule network (both with and without the reconstruction network) as well as 100 from the baseline CNN. The images are stacked and ordered in a 10x10 grid by decreasing activation value; thus, the top left image has the highest activation value of the 100 trials while the bottom right image has the lowest value. Recall that the activation maximization images represent the aggregation of features that the network has learned to represent the particular class. From the images in Fig. \ref{fig:am_6_caps}, we can see that the visualizations are very indicative of the class in question. All of the images show the defining characteristics of a "6" digit; that is, both the circular loop at the bottom as well as the upward curving tail. This shows that the capsule network has learned these facets of the class and use all of them when performing classification. 

This is in contrast to the activation maximization images from the CNN, as shown in Fig. \ref{fig:am_6_cnn}. In this case, the features that the network makes use of are much less clear. Generally, it can be seen that the CNN has some general oblong shapes in the lower half of the images that are likely related to the circular loop of a "6". That said, the clarity of these loops are far worse than those of the capsule network. This supports the notion that the CNN is only searching for an oblong loop in the bottom of the image to classify a "6". Given a CNN's proclivity to find the lowest complexity feature required to discriminate between classes \cite{punjabi2017visualization}, this phenomenon is fairly logical. That said, it is consequently important to look at this distinction in the context of the differences between the capsule network and the CNN. Both networks perform the task extremely well, as each of them obtains over 99\% classification accuracy. Thus, the differences in the visualizations can be attributed to feature complexity rather than just classification power. Therefore, there are indications that the capsule network features capture more information from a class than its CNN counterparts. Fundamentally, the capsule features demonstrate an understanding of the class exemplars to a much greater extent than a CNN feature, which is aligned with the ideas in \cite{sabour2017dynamic}.

The images made from the capsule network without reconstruction further paint the picture of how the capsule features appear. In \ref{fig:am_6_caps_norecon}, one can see that the visualizations are much less interpretable than in the strong reconstruction case. While the features are potentially visible, they are masked by a large amount of noise. Given that the reconstruction places emphasis on training a network to have the ability to invert features back into the original image, it is natural that the capsule network without reconstruction would have much less interpretable features. Furthermore, the distinction is in line with the classification error rates in Table \ref{table:classification}, where the capsule network without reconstruction performs worse.

When looking at other classes, one sees similar results. In Fig. \ref{fig:am_1}, the same procedure is performed on the "1" digit class. The capsule network visualizations have a few more artifacts than in the "6" case, as one can see in the last row, but the general trend still holds. The capsule features are much more descriptive and representative of the members of the dataset class whereas the CNN features are very minimalist. In this particular case, we see that the CNN features are primarily vertically oriented lines. This is, of course, logical for the class in question, but it also neglects the potential distinctions between members within the class. Some "1" images are just vertical lines, others may include a vectored top, and others still have the horizontal underline. Here, the visualizations indicate that the capsules are able to codify intra-class variability to a greater extent than CNN features, which again follows the rationale of \cite{sabour2017dynamic}.

\subsection{Energy Compaction}

\begin{figure*}[ht]
\centering
\includegraphics[width=\linewidth]{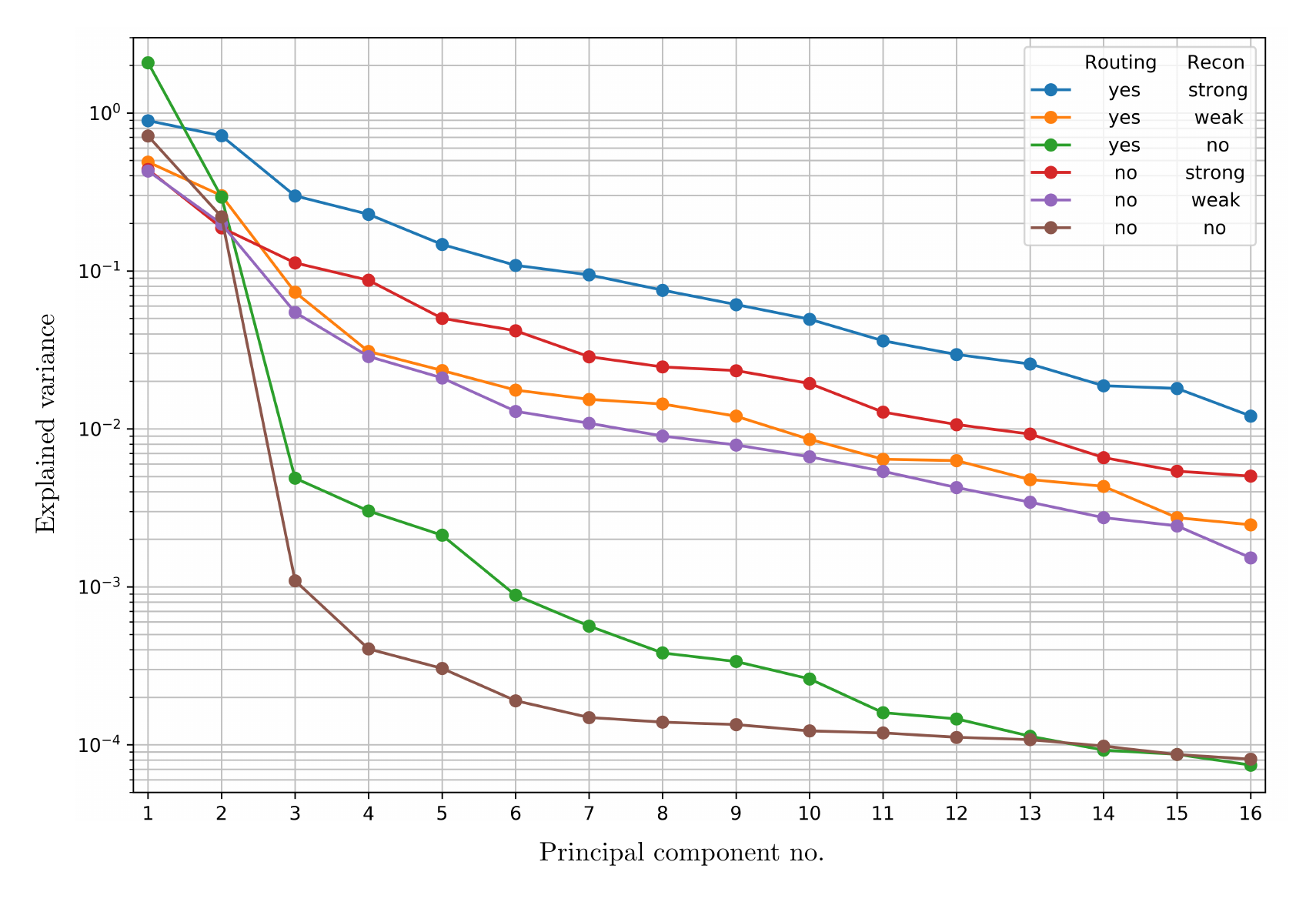}
\caption{Energy compaction analysis on different network configurations}
\label{fig:energy_compaction_results}
\end{figure*}

The activation maximization analysis, while enlightening in its own right, is somewhat limited by its qualitative nature. Thus, the described energy compaction analysis provides a quantitative foil to the visualization results. All of the images in the testing set are passed through the trained network and the resulting capsule vectors are extracted. Then, PCA is performed on the set of capsule vectors. From here, the variances for each of the components are calculated and are shown in Fig. \ref{fig:energy_compaction_results}. Each line corresponds to a capsule network that has varying amounts of reconstruction and routing, as was the case in the classification error comparison.

To reiterate, the variance of a particular PCA components corresponds to the relative amount of information that is contained in that vector dimension. Naturally, the variance decreases for each subsequent principal component; however, the rate of decline is indicative of how much the information is spread amongst the capsule dimensions. An effective capsule vector has a gradual variance decline over its components as the information is distributed effectively, while an ineffective vector has the vast majority of information contained in the first few principal components. Looking at Fig. \ref{fig:energy_compaction_results}, we see that the network with the most gradual decline includes routing and strong reconstruction. On the other hand, the network with the sharpest decline is the one without reconstruction or routing. This indicates that the features in the no reconstruction and routing network are not as information rich, which is consistent with the classification error and activation maximization results as well. Furthermore, we can see that regardless of the presence of routing, increasing the weight of reconstruction improves information distribution. In fact, comparing the routing vs. no routing networks for each reconstruction scenario, we see that the curves are relatively close. The difference is non-negligible, as the routing curves are consistently higher than the non-routing curves; yet, the gap between these two curves is small in all reconstruction cases. Moreover, changing the reconstruction strength yields a bigger change in the curves, especially when the reconstruction is removed altogether. This builds on the results seen in the activation maximization analysis and supports the notion that the presence of the reconstruction loss is pivotal to the overall efficacy of the capsule networks. Additionally, this gives credence to the idea that the reconstruction loss is in fact more important to capsule function than the presence of dynamic routing. This point was perhaps only vaguely alluded to in the classification error comparison, but this energy compaction analysis certainly elucidates it. This concept is not as well explored in \cite{sabour2017dynamic}, but is nonetheless extremely important to the understanding and usage of capsules. While dynamic routing does improve the information distribution in capsule features, the reconstruction network is potentially much more important for the desired behavior of the capsules.

\subsection{Transformation Encoding}

\begin{figure*}[ht]
\centering
\begin{subfigure}{.45\textwidth}
    \centering
    \includegraphics[width=\linewidth]{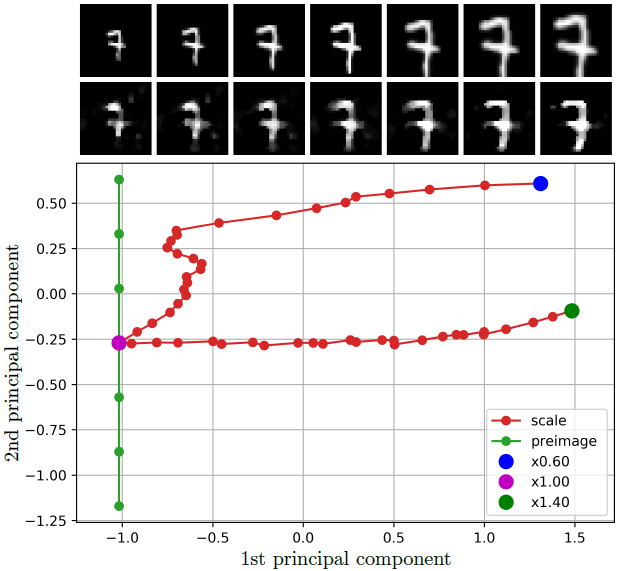}
    \caption{Scale modification}
    \label{fig:trans_scale}
\end{subfigure}
\begin{subfigure}{.45\textwidth}
    \centering
    \includegraphics[width=\linewidth]{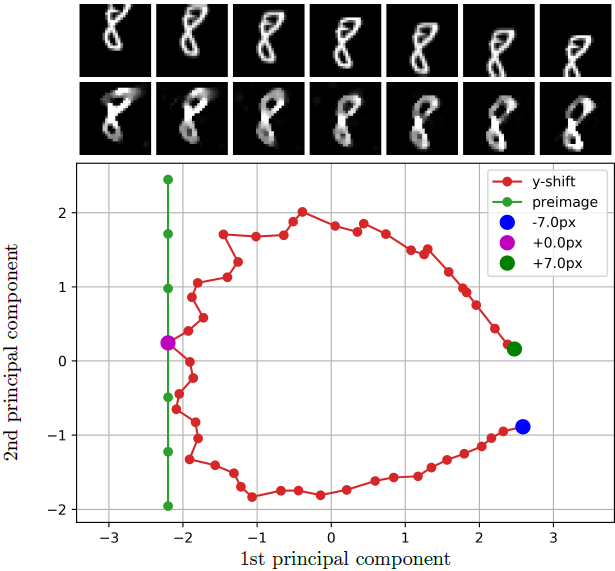}
    \caption{Vertical pixel shift}
    \label{fig:trans_yshift}
\end{subfigure}
\begin{subfigure}{.45\textwidth}
    \centering
    \includegraphics[width=\linewidth]{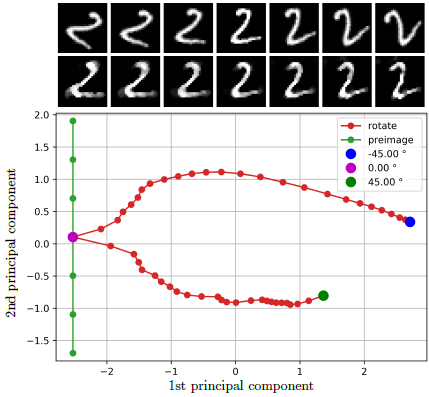}
    \caption{Rotation}
    \label{fig:trans_rotate}
\end{subfigure}
\begin{subfigure}{.45\textwidth}
    \centering
    \includegraphics[width=\linewidth]{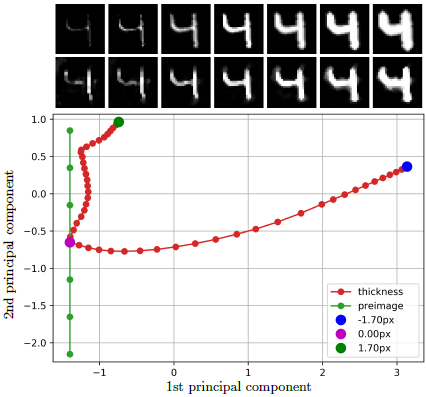}
    \caption{Dilation and erosion}
    \label{fig:trans_thickness}
\end{subfigure}
\caption{Transformation encoding analysis (forward and pre-image) results}
\label{fig:trans_results}
\end{figure*}

To conclude the investigation into capsule features, a transformation encoding analysis was performed in order to examine the ability for the capsule vectors to encode attributes of the application classes. The authors in \cite{sabour2017dynamic} perform a small scale analysis wherein they show that varying a single value in the capsule vector results in specific visual transformations in the reconstructed image. In this manner, the authors explored the impact of the reconstruction network rather than the capsule network itself. For example, modifying one of the capsule values would change the scale of the object in the reconstructed image while another would translate the object. In doing so, the authors claim that this demonstrates that the individual capsule values are able to encode instantiation parameters of class objects, which in turn supports the conclusion that capsules are more robust to these modifications than traditional CNN features. However, this investigation is only a very small (and not thoroughly described) set of experiments. To this end, we performed a more in-depth investigation that consists of both a forward and pre-image analysis. As previously described, the forward analysis takes a set of manually transformed images and performs PCA in order to create a "map" of visual transformations via capsule modifications in the PCA space. Then, a pre-image analysis is done by modifying capsule values in the PCA space and examining visual changes in the image. This is similar to the original experiments in \cite{sabour2017dynamic}; however, in this case the capsule changes are well documented and one can see these changes in the context of distance travelled in the PCA space. Furthermore, because we are showing the changes in the input image space, our results more effectively show the impact of the capsule network rather than the reconstruction network.

In Fig. \ref{fig:trans_results}, one can see four different variations of the transformation encoding experiment. Each case shows the results of both the forward and pre-image analysis. To begin, a single image from the dataset is manually modified by a particular transformation. These images are shown in the top row; for instance, in Fig. \ref{fig:trans_scale}, one can see that the original "7" image (center of top row) is scaled both up and down by up to 40\%. Only a subset of the total number of created images are shown. These images are taken to the PCA space and plotted by the first two principal components along the red curve. This curve spans the manifold on which the image can be scaled up or down. Then, the second principal component is modified to yield points shown on the green line. The second component was chosen experimentally, as modifying the first principal component did not yield the desired transformation. Rather, the objects lost definition to the point where they did not resemble the original class members. This may point towards the fact that the first principal component controls class identity whereas the remaining control the various instantiation parameters. This follows from the construction of the capsules due to the fact that the capsule lengths are used to determine class identity. If the capsule are primarily used for classification, then the length may be the most important facet of the vectors, and therefore the largest variation that the PCA pulled out could have been by lengths. Regardless, these points are then transformed back into the native capsule vector space and are used to create images via the modified activation matching procedure. These image are shown in the second row of each subfigure.

To begin, we can see from Fig. \ref{fig:trans_scale} that the capsules are generally able to reproduce scale changes in the image with small amounts of distortion. Similar patterns can be seen when other transformations, such as y-shift and thickness, are modified. Rotation was found to be a more difficult transformation to emulate, potential because the variability of rotation in the original training set was likely very small. In Fig. \ref{fig:trans_rotate}, one can see from the second row of images that the pre-image analysis was generally only able to slightly rotate small parts of the object. For example, the bottom tail of the "2" is only rotated in the counter-clockwise direction (rightmost images) whereas the top of the "2" remains somewhat stationary. The reverse is true in the clockwise case, where the top part of the object is able to rotate more easily than the bottom. Again, this is likely because the network was not shown images with large rotations during training, so it is unlikely that large rotations would need to be encoded in the capsules.

The distortion in the pre-images generally occurs when the PCA modifications result in capsule vectors that diverge from the original manifold (red curve). This is also quite logical: when the vectors diverge from the original "scale" manifold, it is very likely that other visual changes should occur that may be tangential to simple scaling. In this case, the object lose some of their definition; yet, the quintessential object features remain. This more readily justifies the thesis proposed in \cite{sabour2017dynamic}, as one can see that modifying several capsule values in a manner close to the observed capsule changes in the PCA space via the forward analysis gives images that follow the visual trend. This is a more comprehensive view of the transformation encoding power of capsules than in \cite{sabour2017dynamic}, as the authors there claim that single capsule values can control each facet. This may not necessarily be the case, depending on the model that results from the training procedure. However, even when this is not the case, this analysis shows that ordered modification of multiple components can result in the same phenomenon. Thus, the capsule vectors do indeed encode instantiation parameters and this can be an asset over CNNs in classification tasks.

\section{Conclusion and Future Work}

As shown, capsule network features do fundamentally operate differently than CNN features. In the activation maximization analysis, one could see that the capsule features were better able to describe all facets of the class objects than the CNN features. That said, when the reconstruction network was removed from the capsule pipeline, the features degraded and did not have as much discriminative power. In the energy compaction experiment, we showed that capsules with routing and reconstruction were adept at spreading information across all the elements of the capsule vectors. As the reconstruction weighting was reduced, so too did the information become condensed within one of two principal components, which is more in line with how a scalar CNN feature may behave. Finally, the transformation encoding analysis showed that the capsules are indeed able to capture instantiation parameters of class objects, which is a major benefit over CNN features. The sum of these experiments show that capsule features do have the potential to surpass CNN features, but it is important to note that the reconstruction part of the capsule networks is essential for the desired behavior, whereas the dynamic routing algorithm may not be as beneficial.

To further the work started in this investigation, applying these techniques to a more complicated dataset may produce more discernible differences in classification rates. This may obfuscate the ability to compare features, as the better performing network would naturally have more discriminative features, but there may be benefits to having an experiment where the performances of all networks do not exceed 99\% classification accuracy. Additionally, given how important the reconstruction network was to capsule performance, it may be valuable to compare these results with a CNN that similarly includes a reconstruction network for regularization. Finally, looking at more advanced capsule architectures, such as those with deeper capsule connections or with a different routing scheme, would be valuable. In this manner, one could truly ascertain whether these network are in fact the next stage of evolution in solving computer vision tasks with neural networks.

\section*{Acknowledgment}

The authors would like to thank the Integrated Data Driven Discovery in Earth and Astrophysical Sciences (IDEAS) program at Northwestern University (NSF Research Traineeship Grant 1450006) and the Biomedical Data Driven Discovery (BD3) training program at Northwestern University (NIH Grant 5T32LM012203-02) for financial support.

\bibliographystyle{IEEEtran}
\bibliography{biblio}

\end{document}